\documentclass[sigconf]{acmart}

\AtBeginDocument{%
  }

\setcopyright{acmlicensed}
\copyrightyear{2024}
\acmYear{2024}
\acmDOI{XXXXXXX.XXXXXXX}

\acmConference[KDD 2024 2nd Workshop]{Causal Inference and Machine Learning in Practice}{August 26, 2024}{Barcelona, Spain}

\acmISBN{978-1-4503-XXXX-X/18/06}

\begin{document}

\title[Enhancing Uplift Modeling in Multi-Treatment: Score Ranking and Calibration]{Enhancing Uplift Modeling in Multi-Treatment Marketing Campaigns: Leveraging Score Ranking and Calibration Techniques}

\author{Yoon Tae Park}
\authornote{Both authors contributed equally to this research.}
\email{yoon.park@bestbuy.com}
\affiliation{%
 \institution{Best Buy Co., Inc}
 \country{USA}}

\author{Ting Xu}
\email{ting.xu@bestbuy.com}
\affiliation{%
 \institution{Best Buy Co., Inc}
 \country{USA}}
\authornotemark[1]

\author{Mohamed Anany}
\email{mohamed.anany@bestbuy.com}
\affiliation{%
 \institution{Best Buy Co., Inc}
 \country{USA}}


\begin{abstract}
    Uplift modeling is essential for optimizing marketing strategies by selecting individuals likely to respond positively to specific marketing campaigns. This importance escalates in multi-treatment marketing campaigns, where diverse treatment is available and we may want to assign the customers to treatment that can make the most impact. While there are existing approaches with convenient frameworks like Causalml, there are potential spaces to enhance the effect of uplift modeling in multi treatment cases. This paper introduces a novel approach to uplift modeling in multi-treatment campaigns, leveraging score ranking and calibration techniques to improve overall performance of the marketing campaign. We review existing uplift models, including Meta Learner frameworks (S, T, X), and their application in real-world scenarios. Additionally, we delve into insights from multi-treatment studies to highlight the complexities and potential advancements in the field. Our methodology incorporates Meta-Learner calibration and a scoring rank-based offer selection strategy. Extensive experiment results with real-world datasets demonstrate the practical benefits and superior performance of our approach. The findings underscore the critical role of integrating score ranking and calibration techniques in refining the performance and reliability of uplift predictions, thereby advancing predictive modeling in marketing analytics and providing actionable insights for practitioners seeking to optimize their campaign strategies.
\end{abstract}




\maketitle

\section{Introduction}
Marketing campaigns are continuously seeking optimization to identify the right customers who will be most impacted by specific interventions. Uplift modeling\cite{pmlr-v67-gutierrez17a, K_nzel_2019, zhao2020uplift} has emerged as a powerful tool to address this need by identifying the causal effect of a campaign on individual outcomes. Specifically, uplift modeling quantifies this causal effect using the Conditional Average Treatment Effect (CATE), which has proven effective in various scenarios. However, real-world marketing campaigns can be complex, often involving multiple treatment methods within a single campaign. In these multi-treatment marketing campaign scenarios, more detailed and sophisticated approaches are required.

Several libraries, such as CausalML\cite{chen2020causalml} or EconML\cite{econml}, offer convenient ways to apply uplift modeling in multi-treatment cases. However, a question arises: 'Why not use multiple single-treatment models, applying single treatment modeling for each treatment and then selecting the highest effect treatment for each individual?' This approach can be particularly useful when there are many features to consider, allowing for feature selection tailored to each treatment. While this method can serve as an alternative to existing multi-treatment solutions, it necessitates additional steps, such as score ranking and calibration techniques, to enhance the reliability of uplift predictions.

In this paper, we introduce a novel approach to optimizing multi-treatment marketing campaigns based on CausalML. Instead of directly applying the multi-treatment logic provided by existing framework, we propose using multiple single-treatment models, augmented by score ranking to directly compare treatment effects on individuals, and calibration techniques to improve the reliability of uplift predictions.

Our main contributions are as follows:
\begin{itemize}
\item Applying multiple single-treatment models and selecting the best treatment for each individual, rather than relying on a single multi-treatment model
\item Implementing a score ranking method to ensure that treatment effects are on the same scale for accurate comparison
\item Utilizing calibration techniques to enhance the robustness of uplift models
\end{itemize}

These contributions aim to advance the precision and reliability of uplift predictions, thereby providing more actionable insights for optimizing marketing campaign strategies. Through extensive experiments with real-world datasets, we demonstrate the practical benefits and superior performance of our approach, underscoring the critical role of integrating score ranking and calibration techniques in predictive modeling for marketing analytics.

\section{Related work}
Uplift modeling in multi-treatment contexts has been approached in various ways.

\subsection{Decision tree-based approach} Decision tree-based approaches provide logics in terms of splitting criteria to effectively handle multi-treatment cases. One such approach defines the split criteria based on arbitrary divergence between two distributions: (1) all treatments and the control group, and (2) the differences between treatments themselves \cite{cite-key}. Another decision tree-based approach, the Contextual Treatment Selection (CTS) algorithm, directly maximizes the expected response from the decision tree by optimizing the splitting criteria \cite{zhao2017uplift}.

\subsection{Meta-learner based approach} Extending meta-learners\cite{K_nzel_2019} to support multiple treatments has been proposed by expanding the propensity score model to calculate the Conditional Average Treatment Effect (CATE) for each treatment group \cite{zhao2020uplift}. This approach also introduces a net value optimization framework by modifying treatment effect. The lift is measured using the observed value, the estimated response, and considering the cost function for each treatment.

\subsection{Multi-armed bandit approach}Another interesting work utilizes a multi-armed bandit approach to define customer-level treatment effects by measuring the difference between receiving treatment $t$ and not receiving treatment $t$, where $t$ refers to one of the multiple treatments \cite{sawant2018contextual}. This method follows a logic similar to one-vs-all multi-treatment classification. 

\subsection{Optimization framework} An optimization framework has also been proposed as a direction for improvement. While it primarily relies on single-treatment cases, a dynamic calibration technique \cite{Goldenberg_2020} has been introduced. This technique adjusts the model threshold dynamically and optimizes the incremental treatment outcome, ensuring compliance with the required Return on Investment (ROI) constraints. 


\section{Methodology}
\subsection{Problem Definition}
Our problem focuses on recommending a group of customers who would benefit most from the marketing campaign. Given a limited budget, the marketing team aims to maximize the campaign's impact.

Our uplift modeling framework is based on Rubin's approach \cite{rubin1974estimating}. We denote \( Y \) as the response flag (or revenue) for customer \( i \) at treatment \( t \) (\( t=0 \) for the control group, \( t=1 \) for the treatment group), \( X \) as a vector of features, and \( x_i \) as the feature value per customer \cite{zhao2020uplift}. Here, we rank customers by the highest lift, defined as having the highest Conditional Average Treatment Effect (CATE), where:

\begin{equation}
\tau(x_i) = \mathbb{E}[Y_i(1) - Y_i(0) \mid X = x_i]
\end{equation}

For the multi-treatment case, we expand this formula as follows:

\begin{equation}
\tau_t(x_i) = \mathbb{E}[Y_i(t) - Y_i(0) \mid X = x_i]
\end{equation}

where \( t = n \) (\( n > 0 \)) denotes a treatment condition \( n \), and \( t = 0 \) denotes the control condition, assuming we have \( n \) treatments in the given marketing campaign.

We then compare each treatment effect and select the most effective treatment for each customer as:

\begin{equation}
\mathop{\mathrm{assign\;treatment}}_{t_i} = \mathop{\mathrm{max}}_{\{t_i\}} \tau_{t_i}(x_i)
\end{equation}

Finally, we establish a cutoff point (or threshold) to assign customers to treatment groups based on the top percentile, where we can achieve the highest lift. 

\subsection{Meta-Learner as a baseline}
Our baseline approach is mainly focusing meta-learner, as it has been constantly developed and now is one of the robust approach. This ranges from simple approach such as S/T learners to more complex approach such as X learner. Note that we used those learners from CausalML\cite{chen2020causalml} package for convenience.

\subsubsection{S-learner}
"Single" estimator learner\cite{K_nzel_2019, chen2020causalml} would be considered as one of the simplest approach in meta-learner. S-learner estimates the treatment effect using a single machine learning model by estimating the average outcome $\mu(x)$ with covariates $X$ and an indicator variable for treatment $T$:
\begin{equation}
\mu(x, t) = E[Y \mid X = x, T=t]
\end{equation}
and define the CATE estimate as:
\begin{equation}
\hat{\tau}(x) = \hat{\mu}(x, 1) - \hat{\mu}(x, 0)
\end{equation}

\subsubsection{T-learner}
"Two" estimator learner \cite{K_nzel_2019, chen2020causalml} is probably one of the well-known approaches in meta-learner. This "Two Model Approach" takes two steps. First, it estimates the control response function by a base learner as:
\begin{equation}
\mu_0(x) = E[Y(0) \mid X = x]
\end{equation}
Second, it estimates the treatment response function as:
\begin{equation}
\mu_1(x) = E[Y(1) \mid X = x]
\end{equation}
with a potentially different base learner.

Finally, it estimates the treatment effect by the differences between treatment and control response functions as:
\begin{equation}
\hat{\tau}(x) = \hat{\mu}_1(x) - \hat{\mu}_0(x)
\end{equation}

\subsubsection{X-learner}
The X-learner builds on the T-learner but differs in that it uses the ground truth observations in the training set in an "X"-like shape \cite{K_nzel_2019}. This approach is well-known for using "pseudo-effects". First, it uses the same treatment and control response functions as:
\begin{equation}
\mu_0(x) = E[Y(0) \mid X = x]
\end{equation}
\begin{equation}
\mu_1(x) = E[Y(1) \mid X = x]
\end{equation}

The treatment effect is then estimated by imputing the user level treatment effects $D_i^1$ and $D_j^0$ for user $i$ in the treatment group based on $\mu_0(x)$, and user $j$ in the control groups based on $\mu_1(x)$:
\begin{align}
D_i^1 &= Y_i^1 - \hat{\mu}_0(X_i^1) \\
D_j^0 &= \hat{\mu}_1(X_j^0) - Y_j^0
\end{align}

Using the user level treatment effects $D_i^1$ and $D_j^0$, it estimates $\tau_1(x) = E[D^1 \mid X = x]$ and $\tau_0(x) = E[D^0 \mid X = x]$. Finally, CATE estimate is defined by a weighted average of $\tau_1(x)$ and $\tau_0(x)$ as:
\begin{equation}
\tau(x) = g(x)\tau_0(x) + (1 - g(x))\tau_1(x)
\end{equation}
where $g \in [0, 1]$.

\subsubsection{Meta-learners in Multi-Treatment Cases}
As a baseline approach, we follow methodologies for multi-treatment cases proposed and embedded in \texttt{CausalML} \cite{zhao2020uplift, chen2020causalml}. This approach utilizes propensity scores to estimate the CATE, and handle cost differences across treatments by updating the cost function in the treatment effect formula accordingly.

\subsection{Calibration with Meta-Learner}
A classifer is well-calibrated if the class probabilities accurately represent the true probabilities. For uplift modeling, we extend it to calling the model well calibrated if given different priors, treatments or control, the predicted outcome is close to the real outcome on individual level.  In the context of uplift modeling, calibration ensures that the probability estimates produced by our models are accurate and reliable, thereby improving decision-making processes in marketing campaigns. The methodologies behind calibration involve adjusting the model outputs to correct any biases or inaccuracies, making the predicted probabilities more reflective of true likelihoods.

\subsubsection{Calibration Techniques}
Isotonic regression is a non-parametric method that fits a non-decreasing function to the predicted probabilities. This approach is beneficial when the predicted probabilities do not exhibit a monotonic relationship with the true probabilities, as it ensures a more accurate alignment.

\subsubsection{Implementation}
Utilizing the \texttt{CalibratedClassifierCV} module from \texttt{sklearn\cite{scikit-learn}}, we applied isotonic regression to ensure that our uplift models produce well-calibrated probability estimates. The implementation process involves:

\begin{itemize}
    \item Training uplift models using Meta-Learner frameworks (S-learner, T-learner, X-learner)\cite{K_nzel_2019}
    \item Applying isotonic regression\cite{10.5555/645530.655658, 10.1145/775047.775151} to the model outputs.
    \item Evaluating the calibrated models using Area Under the Uplift Curve(AUUC)\cite{pmlr-v67-gutierrez17a, cite-key}, uplift score plotting, and accuracy on the validation dataset.
\end{itemize}
Calibration enhances the robustness and reliability of our predictions, ensuring that the uplift scores reflect true probabilities of treatment effects. This improves the accuracy of targeting and decision-making in marketing campaigns.

\subsubsection{Measuring Calibration} To assess the effectiveness of calibration, we compared uplift modeling with and without calibration using the AUUC score. The process involved:

\begin{itemize}
    \item \textbf{AUUC Score Comparison:} The metric is calculated by sorting the individual on their predicted uplift score from highest to lowest, divided the sorted group into 100 groups, and computing the difference between treatment and control across these segments. We measured the AUUC for models before and after applying calibration techniques.
    \item \textbf{Uplift Score Distribution:} Plot the uplift curve by placing the cumulative uplift on the y-axis and the proportion of the population on the x-axis. By observing the distribution of uplift scores, we compared models before and after applying calibration techniques. 
    \item \textbf{Uplift at Different Quantiles:} Sorting individual based on their uplift score from highest to lowest. We compared the conversion rates among the top 10\%, 20\%, and other quantiles between the calibrated and uncalibrated models. 
\end{itemize}

\subsection{Scoring Rank: Offer Selection Strategy}
In multi-treatment marketing campaigns, selecting the most effective treatment for each individual is a complex task. To address this, we propose a scoring methodology which involves standardizing uplift scores to identify the best offer. Furthermore, we conduct a comparative analysis between this standardization approach and the direct ranking of uplift scores to evaluate the effectiveness in selecting the most suitable treatment.

\subsubsection{Ranking the Uplift Score and Comparing Ranks}
Calculate uplift scores for each treatment using Meta-Learner models. Rank individuals based on their uplift scores for each treatment and compare these ranks to identify the treatment with the highest uplift for each individual. Ranking individuals based on their uplift scores helps in identifying the most responsive candidates for each treatment, thereby optimizing the allocation of marketing resources.

\subsubsection{Z-score Standardization of Uplift Scores}
Compute the mean and standard deviation of the uplift scores for each treatment. Transform the uplift scores into Z-scores and compare the Z-scores across treatments for each individual. Based on the comparison, select the optimal treatment. Standardizing uplift scores using Z-scores allows for comparisons on a common scale, it can highlight the most impact treatments.
\\
After applying either the direct ranking method or Z-score standardization, integrate the results to select the optimal treatment for each individual. We validated the outcomes using cross-validation and out-of-sample testing. This approach ensured the robustness of the methods and allowed us to determine which technique performed better.

\section{Experiment Setup}
The experimental setup involved utilizing vehicle A, and the campaign was designed as a follow-up to a prior marketing campaign conducted in recent year. This section outlines the details of the campaign, and A/B test experiment setup. 

\subsection{Campaign Overview}

\subsubsection{Initial Marketing Campaign:}
\begin{itemize}
    \item \textbf{Objective:} The initial campaign targeted customers who had received a catalog but had not made a purchase since receiving it.
    \item \textbf{Timeframe:} The catalog purchase window was in short-term period. During this observational period,  a group of customers (10\% of the total targeted customers) received a treatment at the beginning and the offer would exipred at the end of the period.
\end{itemize}

\subsubsection{Follow-Up Campaign:}
\begin{itemize}
\item \textbf{Timeframe:} The follow-up campaign was conducted within a brief period following the evaluation of the initial marketing campaign's results.
\item \textbf{Total Circulation:} A large group of customers were targeted in this campaign, divided into two creative versions:
\begin{itemize}
\item \textbf{Version A:} Treatment A was applied, targeting a subset of the overall targeted customers. This group was equally divided between those who had and had not received prior treatment in the initial campaign.
\item \textbf{Version B:} Treatment B was a more generic version, targeting a subset the overall targeted customers. This group included a mix of those who had and had not received prior treatment.
\end{itemize}
Note that both versions contained a sufficient number of customers, and balancing criteria were applied to ensure that both groups are unbiased.
\end{itemize}

\subsection{Uplift Modeling}
To effectively target customers for the follow-up campaign, we leveraged these uplift modeling techniques to identify those most likely to respond to specific actions. The modeling process is detailed as follows:

\begin{itemize}
    \item \textbf{Data Source:} The dataset for model training was derived from one historical Marketing campaign, which had an identical setup to the new campaign and its follow-up campaign.
    \item \textbf{Feature Selection and Model Training:} Features such as customer demographics, past purchase behavior, and marketing communication history were selected for training the uplift models. These features were chosen based on their predictive relevance to customer responses.
\end{itemize}

\subsection{Evaluation Metrics}
To measure the performance of our uplift models and the campaign outcomes, we used several key metrics: Uplift at Different Quantiles, AUUC, and Accuracy.
\begin{itemize}
    \item \textbf{Uplift at Different Quantiles:} Measure the effectiveness of targeting different segments of the customer base.
    \item \textbf{Area Under the Uplift Curve (AUUC):} Evaluate the overall performance of the uplift model.
    \item \textbf{Accuracy:} Determine the correctness of the predictions.
\end{itemize}

\subsection{Experimental Procedure}

\begin{itemize}
    \item \textbf{Data Preprocessing:} The historical campaign data were cleaned and preprocessed, including handling missing values, outliers, and encoding categorical variables. Additionally, select relevant features using statistical analysis and predictive relevance.
    \item \textbf{Model Training:} Uplift models were trained using Meta-Learner frameworks (S-learner, T-learner, X-learner). Calibration techniques, isotonic regression was applied to refine the model outputs.
    \item \textbf{Score Ranking and Standardization:} Uplift scores for each treatment were calculated, followed by directly ranking or standardization using Z-scores for each treatment.
    \item \textbf{Treatment Selection:} Optimal treatments for each individual were selected based on either the direct ranking method or Z-score standardization. Cross-validation and out-of-sample testing were used to ensure robustness.
    \item \textbf{Performance Evaluation:} The models' performances were evaluated using the specified metrics (AUUC, Uplift At Quantiles), comparing our approach against baseline models in sample.
\end{itemize}

\section{Results}

The results of our experimental setup demonstrate the effectiveness of the proposed uplift modeling approach, particularly in the context of multi-treatment marketing campaigns. Our methodology, which integrates score ranking and calibration techniques, showed significant improvements in prediction accuracy and overall campaign performance.

\subsection{Calibration Effectiveness}
Calibration techniques, particularly isotonic regression, significantly enhanced the reliability of the predicted probabilities. Figure shows the calibration tables for the Meta-Learner models before and after calibration. The calibrated model's probabilities align more closely with the true probabilities, demonstrating the effectiveness of the calibration process.

\subsubsection{AUUC Score Comparison}

The Area Under the Uplift Curve (AUUC) was utilized to evaluate the performance of the uplift models. We observed that our proposed approach, which included score ranking and calibration, achieved higher AUUC scores compared to the baseline models. Table~\ref{tab1} summarizes the AUUC scores for different models:

\begin{table}[h]
\centering
\caption{AUUC Score Comparison}
\begin{tabular}{lp{2cm}p{2cm}p{1cm}}
\toprule
\textbf{Model} & \textbf{AUUC\newline(Uncalibrated)} & \textbf{AUUC\newline(Calibrated)} & \textbf{Diff (\%)} \\
\midrule
S-Learner & 0.40 & 0.65 & 63.4 \\
T-Learner & 0.92 & 1.13 & 22.2 \\
X-Learner & 0.92 & 0.62 & -32.6 \\
Random & 0.54 & 0.54 & - \\
\bottomrule
\end{tabular}
\label{tab1}
\end{table}

Calibration enhanced AUUC scores for the S-learner and T-learner, ultimately identifying the highest-performing learner. However, while the decision to select the calibrated learner remains unchanged, the X-learner exhibited a decrease in AUUC score, highlighting the need to not solely rely on calibration steps and remain open to further studies

\subsubsection{Uplift at Different Quantiles}

We also assessed the effectiveness of targeting different customer segments based on uplift scores. Table~\ref{tab2} illustrates the lift at different quantiles for both calibrated and uncalibrated models. The calibrated models consistently outperformed the uncalibrated models across all quantiles, with the most significant improvements observed in the top 10\% and 20\% segments. Also, note that X Learners underperformed aligning the performance on AUUC scores. The conversion rate of the random target group was used as the baseline for comparison.

\begin{table}[h]
\centering
\caption{Lift Comparison on Top 10\% and 20\% quantile}
\begin{tabular}{lccc}
\toprule
 & Uncalibrated & Calibrated & Diff(\%) \\
\midrule
S Learner - Top 10\% & 0.9 & 1.3 & 48.2 \\
S Learner - Top 20\% & 0.9 & 1.1 & 18.6 \\
T Learner - Top 10\% & 2.0 & 2.2 & 6.5 \\
T Learner - Top 20\% & 1.5 & 1.6 & 5.2 \\
X Learner - Top 10\% & 2.1 & 2.0 & -3.7 \\
X Learner - Top 20\% & 1.6 & 1.5 & -4.8 \\
Random & 1.0 & 1.0 & - \\
\bottomrule
\end{tabular}
\label{tab2}
\end{table}

\subsection{Scoring Rank and Offer Selection Strategy}

Our score ranking strategy, involving both direct ranking and Z-score standardization, was pivotal in selecting the optimal treatment for each individual. Tables \ref{tab3} and \ref{tab4} show the distribution of selected treatments for a subset of the population, demonstrating the diversity and precision of our approach.

\subsubsection{Direct Ranking} Table \ref{tab3} presents the baseline approach, where we rank customers from each treatment by it's uplift score and assign customers how has higher ranking between both treatments. This method shows a 7.5\% lift for treatment A. However, for treatment B, there is a -0.8\% lift, raising questions about the validity of comparing treatments directly by uplift score ranking. It is possible that the assignment strategy favored treatment A for customers who showed positive uplift scores for both treatments A and B. We used the conversion rate of random target group and received no treatment as baseline. 
\begin{table}[h]
\centering
\caption{Direct Rank Comparison}
\begin{tabular}{lccccc}
\toprule
& \textbf{Actual A} & \textbf{Actual B} & \textbf{Actual Ctrl} & \textbf{Lift(\%)} \\
\midrule
Offer A & 2.06 & 1.93 & 1.91 & 7.5 \\
Offer B & 2.90 & 2.77 & 2.79 & -0.8 \\
Offer Ctrl & 1.03 & 1.01 & 1.00 & - \\
\bottomrule
\end{tabular}
\label{tab3}
\end{table}

\subsubsection{Z-Score normalization} Table \ref{tab4} illustrates the results after applying the Z-score normalization. Here, both treatments show a positive lift, with treatment A achieving an 8.2\% lift and treatment B achieving a 3.2\% lift. This approach outperforms the baseline method, indicating that Z-score normalization allows for a more balanced and effective comparison between treatments.

\begin{table}[h]
\centering
\caption{Proposed Approach: Z-score Normalization}
\begin{tabular}{lccccc}
\toprule
& \textbf{Actual A} & \textbf{Actual B} & \textbf{Actual Ctrl} & \textbf{Lift(\%)} \\
\midrule
Offer A & 2.78 & 2.56 & 2.57 & 8.2 \\
Offer B & 2.47 & 2.43 & 2.39 & 3.2 \\
Offer Ctrl & 1.04 & 1.01 & 1.00 & - \\
\bottomrule
\end{tabular}
\label{tab4}
\end{table}

The results indicate that our method effectively identifies the most responsive individuals for each treatment, leading to more targeted and efficient marketing interventions. This demonstrates the benefit of using Z-score standardization over direct ranking for treatment assignment, as it provides a clearer understanding of the relative performance of each treatment.

\begin{table}[h]
\centering
\caption{Online experiment result}
\begin{tabular}{lccc}
\toprule
 & Treatment A & Control & Lift(\%) \\
\midrule
Offer A Persuadable & 1.12 & 1.00 & 12.0 \\
\bottomrule
\end{tabular}
\label{tab5}
\end{table}

In addition to modeling and evaluating the offline experiment results, we also leveraged uplift modeling techniques to score customers and select the optimal offer for them in an online experiment. We collected the online experiment results by running an A/B test. For customers predicted to be more responsive to Offer A, we provided Treatment A to 2/3 of them, while 1/3 received no treatment. We observed a 12\% lift in the online experiment, which outperformed the 8.2\% lift observed in the offline experiment data modeling results, as shown in Table \ref{tab5}.

\section{Conclusion}
In this paper, we have presented a novel approach to uplift modeling in multi-treatment marketing campaigns, discussed the integration of score ranking and calibration techniques to enhance prediction accuracy. By using Meta-Learner frameworks with calibration and implementing a scoring rank-based offer selection strategy, we aimed to address the complexities and improve the incremental effect and reliability of uplift predictions.

Our extensive experiments with real-world datasets showed that the proposed methodology had a robust model performance. The findings underscore the effectiveness of score ranking and calibration in refining reliability of uplift predictions. This advancement provides more actionable insights for practitioners seeking to optimize their campaign strategies.

The key contributions and findings of our work include:
\begin{itemize}
\item The implementation of multiple single-treatment models and the selection of the best treatment for each individual, enhancing the granularity and effectiveness of targeting.
\item The development of a score ranking method to ensure comparability of treatment effects across individuals.
\item The application of calibration techniques to improve the robustness of uplift models, ensuring that predicted probabilities are accurate and reliable.
\end{itemize}

While our results are promising, the study does have some limitations, such as comparison of other approaches other than meta-learners or potential calibration issues. 

Future research should explore additional calibration techniques, applying our methodologies into different algorithms, and applications in different domains to further validate and enhance our approach. Addressing these limitations will help in improving the model's accuracy and applicability.

In conclusion, our work advances the state-of-the-art in predictive modeling for marketing analytics, providing a robust framework for practitioners to optimize multi-treatment campaigns effectively. We believe these contributions will serve as valuable guidelines for further innovations in uplift modeling and its applications.

\bibliographystyle{ACM-Reference-Format}
\bibliography{sample-base}


\begin{thebibliography}{13}


\ifx \showCODEN    \undefined \def \showCODEN     #1{\unskip}     \fi
\ifx \showDOI      \undefined \def \showDOI       #1{#1}\fi
\ifx \showISBNx    \undefined \def \showISBNx     #1{\unskip}     \fi
\ifx \showISBNxiii \undefined \def \showISBNxiii  #1{\unskip}     \fi
\ifx \showISSN     \undefined \def \showISSN      #1{\unskip}     \fi
\ifx \showLCCN     \undefined \def \showLCCN      #1{\unskip}     \fi
\ifx \shownote     \undefined \def \shownote      #1{#1}          \fi
\ifx \showarticletitle \undefined \def \showarticletitle #1{#1}   \fi
\ifx \showURL      \undefined \def \showURL       {\relax}        \fi
\providecommand\bibfield[2]{#2}
\providecommand\bibinfo[2]{#2}
\providecommand\natexlab[1]{#1}
\providecommand\showeprint[2][]{arXiv:#2}

\bibitem[Battocchi et~al\mbox{.}(2019)]%
        {econml}
\bibfield{author}{\bibinfo{person}{Keith Battocchi}, \bibinfo{person}{Eleanor Dillon}, \bibinfo{person}{Maggie Hei}, \bibinfo{person}{Greg Lewis}, \bibinfo{person}{Paul Oka}, \bibinfo{person}{Miruna Oprescu}, {and} \bibinfo{person}{Vasilis Syrgkanis}.} \bibinfo{year}{2019}\natexlab{}.
\newblock \bibinfo{title}{{EconML}: {A Python Package for ML-Based Heterogeneous Treatment Effects Estimation}}.
\newblock \bibinfo{howpublished}{https://github.com/py-why/EconML}.
\newblock
\newblock
\shownote{Version 0.x}.


\bibitem[Chen et~al\mbox{.}(2020)]%
        {chen2020causalml}
\bibfield{author}{\bibinfo{person}{Huigang Chen}, \bibinfo{person}{Totte Harinen}, \bibinfo{person}{Jeong-Yoon Lee}, \bibinfo{person}{Mike Yung}, {and} \bibinfo{person}{Zhenyu Zhao}.} \bibinfo{year}{2020}\natexlab{}.
\newblock \bibinfo{title}{CausalML: Python Package for Causal Machine Learning}.
\newblock
\newblock
\showeprint[arxiv]{2002.11631}~[cs.CY]


\bibitem[Goldenberg et~al\mbox{.}(2020)]%
        {Goldenberg_2020}
\bibfield{author}{\bibinfo{person}{Dmitri Goldenberg}, \bibinfo{person}{Javier Albert}, \bibinfo{person}{Lucas Bernardi}, {and} \bibinfo{person}{Pablo Estevez}.} \bibinfo{year}{2020}\natexlab{}.
\newblock \showarticletitle{Free Lunch! Retrospective Uplift Modeling for Dynamic Promotions Recommendation within ROI Constraints}. In \bibinfo{booktitle}{\emph{Fourteenth ACM Conference on Recommender Systems}} \emph{(\bibinfo{series}{RecSys ’20})}. \bibinfo{publisher}{ACM}.
\newblock
\urldef\tempurl%
\url{https://doi.org/10.1145/3383313.3412215}
\showDOI{\tempurl}


\bibitem[Gutierrez and Gérardy(2017)]%
        {pmlr-v67-gutierrez17a}
\bibfield{author}{\bibinfo{person}{Pierre Gutierrez} {and} \bibinfo{person}{Jean-Yves Gérardy}.} \bibinfo{year}{2017}\natexlab{}.
\newblock \showarticletitle{Causal Inference and Uplift Modelling: A Review of the Literature}. In \bibinfo{booktitle}{\emph{Proceedings of The 3rd International Conference on Predictive Applications and APIs}} \emph{(\bibinfo{series}{Proceedings of Machine Learning Research}, Vol.~\bibinfo{volume}{67})}, \bibfield{editor}{\bibinfo{person}{Claire Hardgrove}, \bibinfo{person}{Louis Dorard}, \bibinfo{person}{Keiran Thompson}, {and} \bibinfo{person}{Florian Douetteau}} (Eds.). \bibinfo{publisher}{PMLR}, \bibinfo{pages}{1--13}.
\newblock
\urldef\tempurl%
\url{https://proceedings.mlr.press/v67/gutierrez17a.html}
\showURL{%
\tempurl}


\bibitem[Künzel et~al\mbox{.}(2019)]%
        {K_nzel_2019}
\bibfield{author}{\bibinfo{person}{Sören~R. Künzel}, \bibinfo{person}{Jasjeet~S. Sekhon}, \bibinfo{person}{Peter~J. Bickel}, {and} \bibinfo{person}{Bin Yu}.} \bibinfo{year}{2019}\natexlab{}.
\newblock \showarticletitle{Metalearners for estimating heterogeneous treatment effects using machine learning}.
\newblock \bibinfo{journal}{\emph{Proceedings of the National Academy of Sciences}} \bibinfo{volume}{116}, \bibinfo{number}{10} (\bibinfo{date}{Feb.} \bibinfo{year}{2019}), \bibinfo{pages}{4156–4165}.
\newblock
\showISSN{1091-6490}
\urldef\tempurl%
\url{https://doi.org/10.1073/pnas.1804597116}
\showDOI{\tempurl}


\bibitem[Pedregosa et~al\mbox{.}(2011)]%
        {scikit-learn}
\bibfield{author}{\bibinfo{person}{F. Pedregosa}, \bibinfo{person}{G. Varoquaux}, \bibinfo{person}{A. Gramfort}, \bibinfo{person}{V. Michel}, \bibinfo{person}{B. Thirion}, \bibinfo{person}{O. Grisel}, \bibinfo{person}{M. Blondel}, \bibinfo{person}{P. Prettenhofer}, \bibinfo{person}{R. Weiss}, \bibinfo{person}{V. Dubourg}, \bibinfo{person}{J. Vanderplas}, \bibinfo{person}{A. Passos}, \bibinfo{person}{D. Cournapeau}, \bibinfo{person}{M. Brucher}, \bibinfo{person}{M. Perrot}, {and} \bibinfo{person}{E. Duchesnay}.} \bibinfo{year}{2011}\natexlab{}.
\newblock \showarticletitle{Scikit-learn: Machine Learning in {P}ython}.
\newblock \bibinfo{journal}{\emph{Journal of Machine Learning Research}}  \bibinfo{volume}{12} (\bibinfo{year}{2011}), \bibinfo{pages}{2825--2830}.
\newblock


\bibitem[Rubin(1974)]%
        {rubin1974estimating}
\bibfield{author}{\bibinfo{person}{D.B. Rubin}.} \bibinfo{year}{1974}\natexlab{}.
\newblock \showarticletitle{Estimating causal effects of treatments in randomized and nonrandomized studies}.
\newblock \bibinfo{journal}{\emph{Journal of Educational Psychology}} \bibinfo{volume}{66}, \bibinfo{number}{5} (\bibinfo{year}{1974}), \bibinfo{pages}{688--701}.
\newblock


\bibitem[Rzepakowski and Jaroszewicz(2012)]%
        {cite-key}
\bibfield{author}{\bibinfo{person}{Piotr Rzepakowski} {and} \bibinfo{person}{Szymon Jaroszewicz}.} \bibinfo{year}{2012}\natexlab{}.
\newblock \showarticletitle{Decision trees for uplift modeling with single and multiple treatments}.
\newblock \bibinfo{journal}{\emph{Knowledge and Information Systems}} \bibinfo{volume}{32}, \bibinfo{number}{2} (\bibinfo{year}{2012}), \bibinfo{pages}{303--327}.
\newblock
\showISBNx{0219-3116}
\urldef\tempurl%
\url{https://doi.org/10.1007/s10115-011-0434-0}
\showDOI{\tempurl}


\bibitem[Sawant et~al\mbox{.}(2018)]%
        {sawant2018contextual}
\bibfield{author}{\bibinfo{person}{Neela Sawant}, \bibinfo{person}{Chitti~Babu Namballa}, \bibinfo{person}{Narayanan Sadagopan}, {and} \bibinfo{person}{Houssam Nassif}.} \bibinfo{year}{2018}\natexlab{}.
\newblock \bibinfo{title}{Contextual Multi-Armed Bandits for Causal Marketing}.
\newblock
\newblock
\showeprint[arxiv]{1810.01859}


\bibitem[Zadrozny and Elkan(2001)]%
        {10.5555/645530.655658}
\bibfield{author}{\bibinfo{person}{Bianca Zadrozny} {and} \bibinfo{person}{Charles Elkan}.} \bibinfo{year}{2001}\natexlab{}.
\newblock \showarticletitle{Obtaining calibrated probability estimates from decision trees and naive Bayesian classifiers}. In \bibinfo{booktitle}{\emph{Proceedings of the Eighteenth International Conference on Machine Learning}} \emph{(\bibinfo{series}{ICML '01})}. \bibinfo{publisher}{Morgan Kaufmann Publishers Inc.}, \bibinfo{address}{San Francisco, CA, USA}, \bibinfo{pages}{609–616}.
\newblock
\showISBNx{1558607781}


\bibitem[Zadrozny and Elkan(2002)]%
        {10.1145/775047.775151}
\bibfield{author}{\bibinfo{person}{Bianca Zadrozny} {and} \bibinfo{person}{Charles Elkan}.} \bibinfo{year}{2002}\natexlab{}.
\newblock \showarticletitle{Transforming classifier scores into accurate multiclass probability estimates}. In \bibinfo{booktitle}{\emph{Proceedings of the Eighth ACM SIGKDD International Conference on Knowledge Discovery and Data Mining}} (Edmonton, Alberta, Canada) \emph{(\bibinfo{series}{KDD '02})}. \bibinfo{publisher}{Association for Computing Machinery}, \bibinfo{address}{New York, NY, USA}, \bibinfo{pages}{694–699}.
\newblock
\showISBNx{158113567X}
\urldef\tempurl%
\url{https://doi.org/10.1145/775047.775151}
\showDOI{\tempurl}


\bibitem[Zhao et~al\mbox{.}(2017)]%
        {zhao2017uplift}
\bibfield{author}{\bibinfo{person}{Yan Zhao}, \bibinfo{person}{Xiao Fang}, {and} \bibinfo{person}{David Simchi-Levi}.} \bibinfo{year}{2017}\natexlab{}.
\newblock \bibinfo{title}{Uplift Modeling with Multiple Treatments and General Response Types}.
\newblock
\newblock
\showeprint[arxiv]{1705.08492}


\bibitem[Zhao and Harinen(2020)]%
        {zhao2020uplift}
\bibfield{author}{\bibinfo{person}{Zhenyu Zhao} {and} \bibinfo{person}{Totte Harinen}.} \bibinfo{year}{2020}\natexlab{}.
\newblock \bibinfo{title}{Uplift Modeling for Multiple Treatments with Cost Optimization}.
\newblock
\newblock
\urldef\tempurl%
\url{https://doi.org/10.48550/arXiv.1908.05372}
\showURL{%
\tempurl}


\end{thebibliography}


\end{document}